\ifx\XeTeXversion\undefined\ifx\directlua\undefined\pdfoutput=1 \fi\fi
\documentclass[a4paper,fleqn]{cas-sc}

\usepackage[authoryear,longnamesfirst]{natbib}
\usepackage{pdflscape}
\graphicspath{{figures/}}

\ExplSyntaxOn
\cs_set:Npn \__first_footerline:
  {
    \group_begin:
    \small \sffamily
    \__short_authors: :~
    { \rmfamily \itshape Preprint }
    \group_end:
  }
\ExplSyntaxOff

\begin{document}
\let\WriteBookmarks\relax
\def\floatpagepagefraction{1}
\def\textpagefraction{.001}

\shorttitle{Network-based Spatial Context Retrieval for Open-weight LLMs}
\shortauthors{J. Perez}

\title [mode = title]{Network-based Spatial Context Retrieval for Open-weight LLMs: A Faithfulness Benchmark for Grounded Geographic Reasoning}

\author[1]{Joan Perez}[orcid=0000-0003-3003-0895]
\cormark[1]
\address[1]{Urban Geo Analytics, \href{https://urbangeoanalytics.com}{urbangeoanalytics.com}}
\cortext[1]{Corresponding author. E-mail address: \href{mailto:jperez@urbangeoanalytics.com}{jperez@urbangeoanalytics.com}}

\begin{abstract}
Large language models (LLMs) encode substantial latent geographic knowledge, yet they reason poorly over space and are unreliable when queried from coordinates alone. Useful behaviour emerges only when structured spatial context is supplied in the prompt. This raises a question geographic evaluation has left unexamined: once the right context is supplied, does the model reason from it, or override it with its own parametric recall? We take up this question with an open pipeline for network-based spatial context retrieval. In it, the surroundings of a selected point are defined by the pedestrian street network, the area actually reachable on foot. Using only open data and open-weight models, the pipeline retrieves features from OpenStreetMap and the GHS-POP population grid, computes indicators over the network catchment in code, and injects them as a compact spatial brief. On this basis we build a faithfulness benchmark. It labels every claim a model makes by its source (grounded in the brief, or drawn from training knowledge) and its correctness, and it probes each case with a planted false premise that the brief refutes. We evaluate sixteen open-weight model configurations across three families (Qwen, Gemma and Llama, with Gemma in two generations), four size classes and, where available, both thinking and non-thinking modes, on three contrasting cities, resampling every case over ten seeds. The results show that resistance to the planted premise varies more strongly by model family and generation than by scale, while brief-reading competence forms a partly separate dimension. These behaviours are not captured by conventional world-correctness scores or single-shot evaluation. We release the implementation, spatial briefs, model outputs, and claim-level labels as a reproducible workflow at \href{https://github.com/perezjoan/NSCR-LLM}{github.com/perezjoan/NSCR-LLM}.
\end{abstract}


\maketitle

\section{Introduction}

Large language models (LLMs) encode substantial geographic knowledge, but their performance is unreliable when geographic information is accessed through coordinates alone. \citet{manvi2023} show that models can recover latent information about population density, economic livelihood, and related indicators, while also showing that performance improves substantially when prompts are enriched with structured OpenStreetMap context. At the same time, probing studies find that LLMs struggle with basic spatial operations, including geoparsing, relative spatial relationships, distance, and direction \citep{bhandari2023,yu2025}. Recent benchmarks such as MapEval \citep{dihan2025} likewise report substantial gaps in map-grounded question answering, particularly for distance, direction, and route reasoning.

The resulting picture is therefore mixed: LLMs can interpret geographic information when it is supplied in an appropriate form, but they are unreliable as unaided spatial reasoning engines. This makes the construction of spatial context a central design problem. Rather than asking what a model knows about a location, a more practical question is what information should be retrieved and how it should be presented. But supplying context raises a second question that existing geographic evaluations largely leave unexamined: once the relevant facts are in front of the model, does it actually reason from them? A system that injects current, local data into a prompt is only useful if the model does not override that information with parametric recall or accept a false claim introduced by the user. This faithfulness to supplied context is distinct from correctness about the world, and existing geographic evaluations generally cannot measure it.

This paper addresses both questions through a deliberately narrow setting: given a single location selected on a map, we ask a language model to reason about its immediate surroundings. The approach has three components. First, the surroundings are defined by a pedestrian street-network catchment rather than a Euclidean buffer, so the retrieved area corresponds to what can actually be reached on foot. Second, features from OpenStreetMap and the GHS-POP population grid are aggregated in code into a compact set of indicators and presented to the model as a structured spatial brief, with no fine-tuning or proprietary services. Third, faithfulness is evaluated at the claim level: each model claim is labelled by its source, distinguishing information grounded in the brief from information drawn from training knowledge, and by its correctness. Each case also contains a planted false premise that contradicts the supplied brief, testing whether the model follows the evidence or accepts the assertion.

We evaluate this framework across three contrasting cases (a food-access case in Chicago, a fifteen-minute-city district in Paris, and a compact urban core in Hanoi), sixteen open-weight model configurations spanning Qwen, Gemma, and Llama families, four size classes, and, where available, thinking and non-thinking modes. Each case is resampled over ten seeds. The evaluation examines whether models use the supplied brief, whether they resist a false premise that contradicts it, and how these behaviours vary across models, cases, and generations.

The remainder of the paper reviews related work on language models and geography (Section 2), describes the retrieval pipeline and faithfulness benchmark (Section 3), presents the three cases (Section 4), reports the results (Section 5), and discusses implications, limitations, and future work (Section 6). Section 7 concludes the paper. Appendices A to C provide the indicator definitions, field glossary, and complete per-model results.

\section{Motivation and related work}

\subsection{Language models meet geography}

Four lines of work have emerged, distinguished by where the model's geographic information comes from: recalled from its own weights, produced by its own spatial reasoning, fetched by tools it calls, or supplied as retrieved context in the prompt. The first and the fourth are the ones this paper builds on.

The first route asks what spatial knowledge is latent in the model. GeoLLM \citep{manvi2023} is the clearest statement of this line: LLMs embed enough extractable geospatial information to predict socio-economic indicators such as population density, asset wealth, and mean income, but the knowledge is almost inaccessible through the obvious interface, since prompting with bare latitude and longitude performs poorly. Their method instead builds prompts from auxiliary OpenStreetMap context (the address and nearby places around a coordinate) and fine-tunes on them, yielding a reported 70\% improvement in Pearson's r-squared over coordinate and nearest-neighbour baselines. Two points follow: the effect scales with model size and corpus, so geographic competence tracks scale, and the surfaced knowledge is a compressed summary of the training corpus, and so stale by construction. Complementary probing studies \citep{bhandari2023} reinforce the picture, finding competence uneven across tasks and again correlated with model sophistication. The lesson is that prompt context, not the coordinate, unlocks useful behaviour, and because that latent knowledge is stale, live data is preferable to memory. It is also not neutral: \citet{manvi2024} show LLMs are systematically geographically biased, rating locations in Africa and other under-represented, lower-income regions lower on subjective indicators such as attractiveness, morality, and intelligence even in the zero-shot setting, a prejudiced prior that falls hardest where authoritative open data is already scarcest.

The second route asks how well LLMs reason over space, and the answer is sobering. Models stumble on elementary operations, geoparsing, judging relative spatial relationships, and reasoning over distance and direction \citep{bhandari2023,yu2025}. The most comprehensive assessment, MapEval \citep{dihan2025}, makes the ceiling concrete: across 700 map-based multiple-choice questions spanning 180 cities and 54 countries, and thirty foundation models evaluated over textual, API-based, and visual tasks, no model exceeds 67\% accuracy, open-weight models trail proprietary ones substantially, and every model lags human performance by more than twenty points, with the sharpest failures in distance, direction, and route planning. The implication for design is direct: a system should not delegate spatial computation to a model that cannot reliably sum, compare, or divide, but pre-compute the quantities and leave the model to interpret finished numbers.

The third route is agentic and tool-using: rather than relying on what the model knows or can compute unaided, give it the ability to call GIS operations, spatial databases, or web APIs. The ``Autonomous GIS'' vision of \citet{li2023}, whose LLM-Geo prototype uses the GPT-4 API to chain geoprocessing operations into aggregations, charts, and maps, is the reference point, alongside GeoGPT \citep{zhang2023}, which couples an LLM to a pool of GIS tools, and QGIS-embedded assistants such as GIS Copilot \citep{akinboyewa2025}. It is powerful but shifts complexity into orchestration and depends on heavyweight or proprietary backends, and it re-inherits the second route's spatial-reasoning weaknesses at the planning layer: whenever the model, not a fixed routine, decides what to compute, the documented unreliability re-enters through the choice of operations.

The fourth route neither trusts the model's memory nor asks it to act: it retrieves relevant information at query time and places it in the context window, so that generation is grounded in supplied evidence. This is retrieval-augmented generation (RAG), introduced by \citet{lewis2020} to overcome the limitation that parametric knowledge may be incomplete, outdated, or absent. Unlike tool-use, where the model acts in a loop, RAG assembles the context before generation and leaves the model only to read it. The standard formulation is a poor fit for geography, though, because its notion of relevance is semantic, retrieving the chunks most textually similar to the query, whereas the information most relevant to a place is fixed by spatial proximity and reachability. Spatial-RAG \citep{yu2025} is the most developed response, combining sparse spatial retrieval with dense semantic retrieval under a multi-objective ranking, with clear gains on ``find restaurants along this route'' queries; it already handles points, polylines, and polygons through buffer-like constructs (a radius, a corridor, a boundary), showing that geometry-aware spatial retrieval for LLMs is established.

Closest to our setting is ChatMap \citep{unlu2023}, which takes a circular area around a point (a 300-metre-radius disc in densely tagged districts of Istanbul), assembles the OpenStreetMap features within it into curated prompts, and fine-tunes a small 1B model to answer questions about the location. Framed as groundwork for ``geospatially aware urban RAG,'' it demonstrates our target use-case, a conversational interface to a point's OSM context, and is our most direct predecessor; Section 2.3 states how we depart from it.

\subsection{Faithfulness to supplied context}

Retrieval raises a second question: how does the model use the context it is given? Retrieval helps only if the model grounds its answer in the supplied information rather than overriding it with parametric recall, and if it does not accept false claims introduced through the context or the user. This faithfulness to supplied context is distinct from correctness about the world and is not guaranteed.

Prior work shows that models can ignore context that conflicts with their parametric knowledge, generating information absent from the input \citep{longpre2021,xu2024}. They may persist with a confident but incorrect prior despite correct supplied evidence \citep{longpre2021,jin2024}, or accept fluent external evidence even when it is false \citep{xie2023,pan2023}. \citet{wu2024} describe this as a conflict between retrieved context and internal priors. Providing the right information in the context window therefore does not ensure that the model will use it.

This distinction is particularly important for grounded geographic reasoning. Existing geographic evaluations primarily measure whether an answer is correct about the world. They cannot distinguish a model that correctly uses the supplied brief from one that ignores it but happens to produce the right answer, nor can they identify when a model contradicts supplied data or accepts a false premise. General RAG evaluation has increasingly moved toward claim-level auditing, but to our knowledge this approach has not been applied systematically to faithfulness in geographic reasoning. Our benchmark addresses this gap by labelling individual claims according to both their source and correctness.

A further limitation of existing evaluations is their reliance on single generations. Because language models are stochastic, adherence to supplied context and resistance to a false premise can vary across samples. A single completion therefore captures one realization rather than a model's characteristic behaviour. We instead repeat each case under a fixed protocol across multiple seeds, allowing faithfulness and resistance to a planted false premise to be evaluated as recurring tendencies rather than as properties of a single generation (Section 4.2).

\subsection{The gap and our approach}

The literature reviewed above points to two gaps. First, existing geographic retrieval systems generally define spatial context using Euclidean distance and provide models with raw or lightly structured geographic features. For questions about what is reachable from a location, however, straight-line proximity does not necessarily represent the accessible area. We therefore define context using a network-based catchment and pre-compute a compact set of spatial indicators over that catchment. The resulting spatial brief contains finished values derived from open geographic data, separating deterministic spatial computation from the model's interpretation.

Second, existing geographic evaluations primarily assess whether models produce correct answers about the world, rather than whether they use the context supplied to them. We address this by evaluating faithfulness at the claim level: each claim is classified by its source, distinguishing information grounded in the spatial brief from information supplied through the model's external knowledge, and by its correctness. We additionally introduce a planted false premise to test whether models defend the supplied evidence or accept a plausible but incorrect assertion.

\section{Method}

\subsection{Overview and design principles}

The system turns a single click on a map into a natural-language conversation about that location, and provides the controlled setting in which we then measure how faithfully a model reasons from it. It has two stages (Figure 1). The first is purely deterministic: from the clicked point it computes a network catchment, retrieves the relevant open data over that catchment, and assembles a compact structured object we call the spatial brief. The second is the language model: the brief is injected into its context window together with a role description, and the user then questions it about the place. Three principles govern the design.

The first is separation of computation from interpretation. Because LLMs reason unreliably over space and arithmetic (Section 2.1), every quantity the model might otherwise have to derive is computed in code beforehand, and the model receives finished numbers and is asked only to interpret them in language. This division of labour is what makes faithfulness measurable since every quantitative claim the model makes can be checked directly against a known brief. The second is a network-defined extent. The neighbourhood of the point is delineated by the pedestrian street network, so that every retrieved feature is something genuinely reachable on foot from the point. The third is openness throughout. The retrieved data are open-source, and the language models are open-weight. No commercial geocoding, routing, or basemap service is used, and no model is fine-tuned, so the whole pipeline, and therefore every result in the evaluation, can be reproduced.

\begin{figure}
  \centering
  \includegraphics[width=0.62\linewidth]{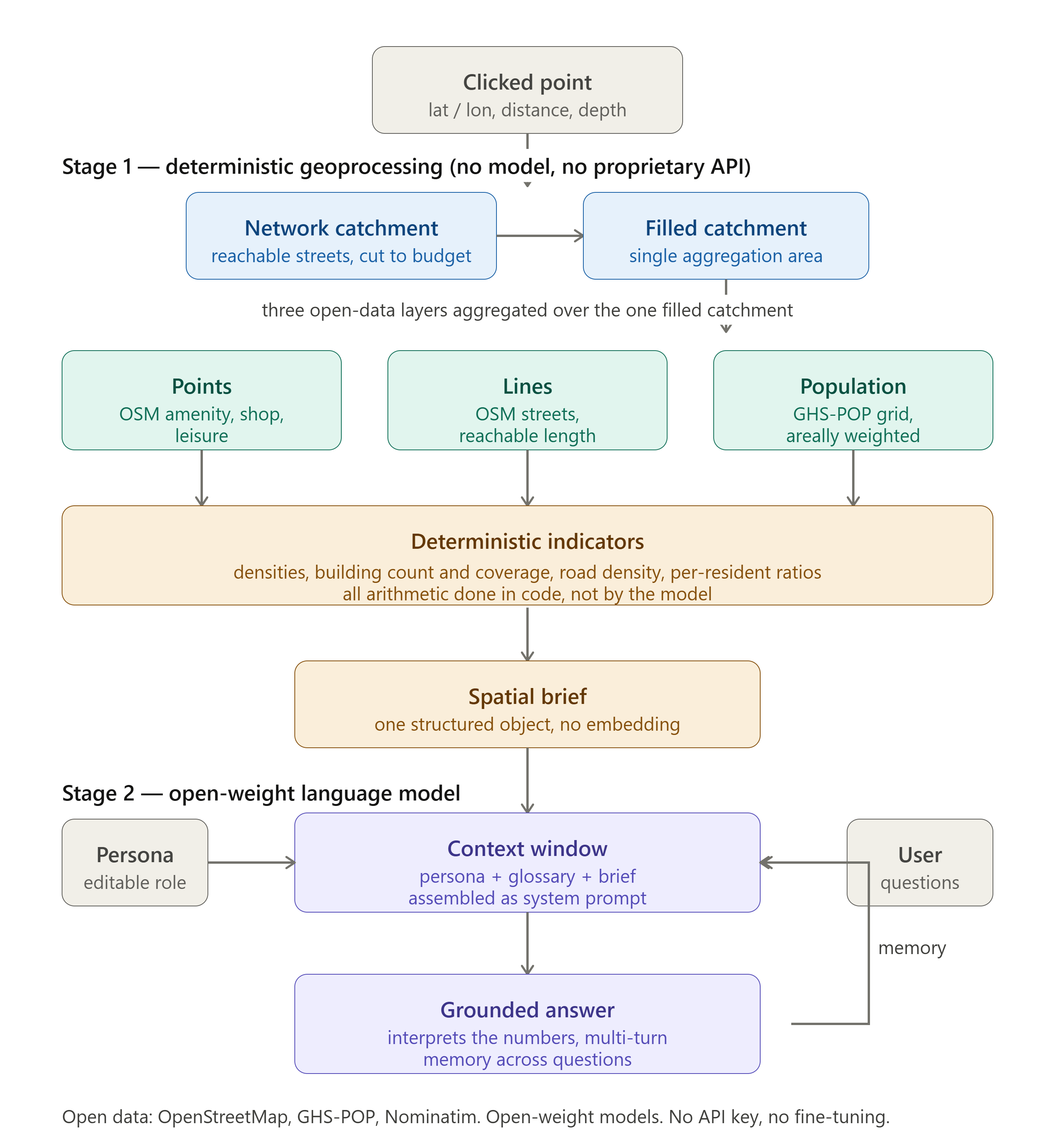}
  \caption{The two-stage pipeline. A clicked point defines a pedestrian network catchment; open data (OpenStreetMap network, POIs, buildings; GHS-POP population) are retrieved and reduced in code to a compact spatial brief, which is injected into an open-weight LLM for questioning.}\label{fig:1}
\end{figure}

\subsection{The network catchment}

Given a clicked point and a walking distance d, the system downloads the surrounding walkable street network from OpenStreetMap using OSMnx \citep{boeing2017} and projects it to a local metric coordinate system, so that all subsequent lengths and areas are computed in true metres. The point is snapped to the nearest network node, and walking distances to every other node are found by shortest-path traversal \citep{hagberg2008}, with segment length as the edge cost, so that reachability follows the streets rather than straight-line distance. The catchment is the set of segments reachable within d. It is edge-based: where only one end of a segment falls within the budget, the segment is cut at the exact remaining distance, so the boundary reflects the true walking limit rather than snapping to intersections. Summing all reachable segments, including cut portions, gives the reachable road length.

These segments form a thin lattice, not an area. To obtain a surface over which the other layers can be aggregated, they are buffered outward by a block-depth parameter b, unioned, and buffered slightly inward again to close gaps between parallel streets. The resulting filled catchment is the single geometry over which population, points of interest, and buildings are all aggregated, which keeps the derived ratios internally coherent. Both parameters, the walking distance d and the block depth b, are recorded in the brief, so the spatial extent behind any set of numbers is always explicit; b sets how far the catchment reaches off the streets and therefore the area over which densities are measured.

\subsection{Retrieving and aggregating the layers}

Once the filled catchment is fixed, it becomes the single spatial query window for every data source. The system retrieves each layer live at the moment of the click, pulling only what falls inside the catchment: features are queried from OpenStreetMap within its boundary, and population is streamed from a remote raster over its bounding box, so nothing is fetched or stored beyond the neighbourhood in question. Three thematic layers are retrieved, one per geometry type, together with a reverse-geocoded place name.

The place name comes from Nominatim, with the region and city fields falling back through several candidate keys (state, region, or province; city, town, village, or municipality) so the name degrades gracefully in smaller settlements rather than returning empty. Residential population is streamed from GHS-POP R2023A \citep{schiavina2023}, a 100 m grid, reading only the pixels over the catchment; each cell is areal-weighted, contributing population in proportion to the share of its area inside the catchment, since a 100 m cell is coarse relative to a small catchment and an all-or-nothing rule would introduce substantial error. The figure is modelled residential population for the 2021 reference year, people who live there rather than daytime workers or visitors, so it can be low even where an area is busy; this caveat is carried into the brief in words, and whether the model respects it is one of the sharpest tests in the benchmark (Section 5). Points of interest (amenities, shops, and leisure features) are queried from OpenStreetMap and retained if a representative interior point falls within the catchment; reducing polygon-mapped features to an interior point rather than discarding them matters, because many substantial destinations such as schools, supermarkets, and parks are mapped as areas, not points. Building footprints, finally, are filtered so that whole buildings are counted rather than fragments: building-parts and footprints below 15 m\textsuperscript{2} are removed, since OpenStreetMap tagging is heterogeneous and cadastral imports often split one structure into many parts. Counts and average footprint use a centroid-in rule, while the coverage ratio separately clips intersecting footprints to the boundary, so it cannot exceed one.

\subsection{The spatial brief}

From the retrieved layers the system computes a small set of indicators in code, covering the catchment's geometry, population, buildings, street network, and points of interest. Ratios are formed by dividing already-aggregated totals rather than averaging per-feature values, so density, for example, is total population over total area. The full indicator set, with the definition and computation rule of each, is given in \textbf{Appendix A}. The result is packaged as a plain structured object with named fields. Alongside the brief, the model receives a short glossary defining every field in plain language; the glossary is part of the fixed prompt, is version-hashed in the run metadata, and is reproduced verbatim in \textbf{Appendix B}. Because the brief is compact and already reduced to finished values, it fits comfortably in a model's context window and leaves no arithmetic for the model to perform; together with the glossary it is the sole channel through which the model learns about the location, which is what lets every claim in an answer be traced unambiguously to the brief or to knowledge from outside it. The brief deliberately departs from conventional retrieval-augmented generation (RAG). Since the underlying data are structured and computed, information is retrieved through deterministic spatial aggregation rather than semantic similarity search. The model therefore receives a computed summary rather than source documents.

\subsection{Injecting the brief into an open-weight model}

In the second stage, the brief becomes the factual context for an open-weight language model. Because the trained parameters are publicly available, these models can be downloaded and run locally without relying on a proprietary API or exchanging data with the provider. The model version is pinned, making the inference process reproducible by anyone using the same weights.

The system prompt is assembled from three parts: a persona that gives the model an analytical role, a short glossary explaining each field of the brief, and the spatial brief itself. Persona and glossary are editable and version-hashed, so the same brief can be presented under different analytical roles while every answer remains traceable to the exact prompt that produced it. The user then questions the model in a multi-turn conversation that retains memory, with each prior exchange replayed into the context on the following turn. Only final answers re-enter this history: where a model emits an explicit reasoning trace before answering, the trace is separated and excluded from subsequent turns, following the manufacturers' own guidance. The model's task is purely interpretive: to translate the finished indicators into grounded statements and sustain a coherent discussion about the place.

The models are loaded locally with four-bit quantisation (NF4 with double quantisation, bfloat16 compute), identical across the grid, so several sizes run on a single commodity GPU and any model can be swapped in without changing the pipeline. We evaluate sixteen configurations drawn from eleven checkpoints across three families and four size classes (\textbf{Table 1}). Qwen3 (Alibaba Cloud) contributes the 1.7B, 4B, 8B, and 14B variants \citep{yang2025}. These hybrids ship a built-in ``thinking'' mode that produces an explicit reasoning trace before the answer and can be disabled in the chat template without touching the weights; we run every size in both modes, giving four same-weights pairs in which the reasoning trace is the only difference. Gemma 4 (Google DeepMind) contributes the dense 12B, the first Gemma generation with an equivalent toggleable thinking mode, likewise run in both modes; its reasoning arrives in a dedicated thought channel that the pipeline parses apart from the answer. Gemma 3 (Google DeepMind) contributes the 1B, 4B, and 12B variants \citep{gemmateam2025}, and Llama 3.x (Meta) contributes Llama 3.2 at 1B and 3B and Llama 3.1 at 8B \citep{grattafiori2024}; neither produces a separable reasoning pass. Gemma 3 accepts no system role, so for that family the system prompt is folded into the first user turn.

Decoding follows each manufacturer's recommended sampling parameters (\textbf{Table 1}), so every model runs in the regime it was tuned for. Where a vendor specifies no value, as Meta does for top-k, the library default is kept rather than an imposed value. This reflects how the models are used in practice, though it means decoding settings differ between families. Because the brief, glossary, persona, and questions are held fixed while only the model configuration varies, the configuration is isolated as the sole source of behavioural difference.

\begin{table}
\caption{The sixteen model configurations evaluated (eleven distinct checkpoints from three families), with thinking-mode setting and the manufacturer-recommended sampling parameters used for decoding.}\label{tab:models}
\begin{tabular*}{\tblwidth}{@{}LCCLLCCCR@{}}
\toprule
Model & T\textsuperscript{a} & Params & Manufacturer & Release & Temp & Top-p & Top-k & Size\textsuperscript{b} \\
\midrule
Qwen3 & Y & 14B & Alibaba Cloud & Apr 2025 & 0.6 & 0.95 & 20 & 29.5 GB \\
Qwen3 & N & 14B & Alibaba Cloud & Apr 2025 & 0.7 & 0.8 & 20 & 29.5 GB \\
Qwen3 & Y & 8B & Alibaba Cloud & Apr 2025 & 0.6 & 0.95 & 20 & 16.4 GB \\
Qwen3 & N & 8B & Alibaba Cloud & Apr 2025 & 0.7 & 0.8 & 20 & 16.4 GB \\
Qwen3 & Y & 4B & Alibaba Cloud & Apr 2025 & 0.6 & 0.95 & 20 & 8.05 GB \\
Qwen3 & N & 4B & Alibaba Cloud & Apr 2025 & 0.7 & 0.8 & 20 & 8.05 GB \\
Qwen3 & Y & 1.7B & Alibaba Cloud & Apr 2025 & 0.6 & 0.95 & 20 & 4.0 GB \\
Qwen3 & N & 1.7B & Alibaba Cloud & Apr 2025 & 0.7 & 0.8 & 20 & 4.0 GB \\
Gemma-4 & Y & 12B & Google & Jun 2026 & 1.0 & 0.95 & 64 & 23.8 GB \\
Gemma-4 & N & 12B & Google & Jun 2026 & 1.0 & 0.95 & 64 & 23.8 GB \\
Gemma-3 & N & 12B & Google & Mar 2025 & 1.0 & 0.95 & 64 & 24.4 GB \\
Gemma-3 & N & 4B & Google & Mar 2025 & 1.0 & 0.95 & 64 & 8.6 GB \\
Gemma-3 & N & 1B & Google & Mar 2025 & 1.0 & 0.95 & 64 & 2.0 GB \\
Llama-3.1 & N & 8B & Meta & Jul 2024 & 0.6 & 0.9 & -- & 16.07 GB \\
Llama-3.2 & N & 3B & Meta & Sep 2024 & 0.6 & 0.9 & -- & 6.43 GB \\
Llama-3.2 & N & 1B & Meta & Sep 2024 & 0.6 & 0.9 & -- & 2.47 GB \\
\bottomrule
\end{tabular*}
\par\vskip3pt
\parbox{\tblwidth}{\footnotesize\raggedright
\textsuperscript{a} T: thinking mode. Y indicates the configuration generates an explicit reasoning trace before its answer; N indicates it answers directly.\par
\textsuperscript{b} Size: the bf16 checkpoint as distributed on Hugging Face. All models are run in four-bit quantisation (NF4).}
\end{table}

\subsection{A faithfulness benchmark}

The benchmark rests on a change of unit. Rather than scoring an answer as a whole, which conflates different types of success and failure, we decompose every answer into atomic claims and score each separately. An atomic claim is the smallest span of text that makes an independently checkable assertion: a number, count, existence statement, characterisation, or conclusion. For example, ``the area is dense, with 8,440 residents and excellent transit'' contains three claims with potentially different outcomes. Answers are therefore exhaustively segmented into bracketed claims using fixed conventions: figures are separated from interpretations derived from them, distinct items in enumerations are split, and collective characterisations are kept whole when the assertion concerns the mix rather than its individual members.

Each claim is then judged along two independent dimensions: \textbf{source} and \textbf{correctness}. The source question is whether the claim could have been produced from the brief and its glossary alone, or whether it introduces information from outside them. Brief-sourced claims are checked against the brief; external claims are checked against the world where possible, with the brief taking precedence if the two conflict. Crossing these dimensions produces four categories: \textbf{brief-true}, for claims that correctly restate, derive from, or fairly characterise the brief; \textbf{brief-false}, for claims that misread or misrepresent it; \textbf{recall}, for correct or reasonably defensible external knowledge; and \textbf{hallucination}, for false, fabricated, unverifiable, or brief-contradicting external content.

A single fabricated passage illustrates all four:

\begin{quote}\itshape
``The catchment houses [8,440 residents](1) at a density of [roughly 27,000 per square kilometre](2), making it [one of the liveliest corners of the Marais](3). Food is well covered, with [over 90 restaurants](4) and [several bakeries](5), and [three metro stations put the rest of the city within easy reach](6).''
\end{quote}

Claims (1), (2), and (5) are brief-true: the first two restate fields, while the third fairly summarises the bakery count. Claim (4) is brief-false, since the brief records 79 restaurants. Claim (3) is recall: the neighbourhood name is absent from the brief but is correct for the coordinates and contradicts nothing. Claim (6) is hallucination: the brief is silent on transit, and the specific station count is unsupported. The example shows why source and correctness must be separated: (4) and (6) are both wrong, but the former reflects a failure to use supplied evidence correctly, while the latter reflects unsupported external content.

Aggregating the labels produces a profile rather than a single score. The share of brief-sourced claims measures how much of an answer is grounded in the supplied context; the error rate among brief-sourced claims measures reading competence; and the split between recall and hallucination measures behaviour when the model goes beyond the brief. These failure modes are therefore measured separately: a model can be a careful reader but an unreliable recaller, or the reverse, patterns that whole-answer scoring cannot distinguish. Section 4 describes how the labels are produced at scale and how the benchmark is applied to the three frozen cases.

\subsection{Implementation and reproducibility}

The system is a single notebook, organised to mirror the two stages of the method. The first cells implement stage one: the interactive map and catchment computation, the population streamer, and the assembly of the structured brief. The remaining cells implement stage two: model loading across the three families, the persona-and-brief system prompt, and the questioning interface, together with a batch runner that replays a fixed question sequence over repeated seeds.

All materials are openly released (\href{https://github.com/perezjoan/NSCR-LLM}{github.com/perezjoan/NSCR-LLM}), with a versioned snapshot archived on Zenodo (doi:\href{https://doi.org/10.5281/zenodo.23056312}{10.5281/zenodo.23056312}), tagged v1.0.0. The archive includes the notebook, persona and glossary used verbatim, environment specification, and the evaluation materials from Section 4: the three site briefs and catchment parameters, raw model outputs, claim-level labels, and the labelling and trap-scoring rubric. It also includes the seed-1 labels produced by the human annotator and language-model judge.

Because OpenStreetMap and GHS-POP evolve over time, the briefs are archived as frozen records rather than redistributing the underlying datasets. Each frozen brief preserves the exact features and indicators presented to the models and used for evaluation. The briefs use GHS-POP R2023A and OpenStreetMap data retrieved on 25 June 2026. The deterministic stage can therefore be rerun against current data to generate an updated brief, while the frozen briefs and released outputs allow the results reported here to be reproduced independently of subsequent changes to the upstream data.\footnote{The deterministic stage is reproducible exactly, but generation is not guaranteed to be bitwise identical across hardware and software configurations. A fixed seed controls sampling within a given configuration, while CUDA-level nondeterminism can produce different completions across GPUs, drivers, or library versions. The released outputs are therefore the authoritative record of the generated responses.}

\section{Application}

\subsection{Three case studies}

We apply the method to three geographically and morphologically distinct locations, each aligned with a well-known strand of geographic literature the models are likely to have encountered during training. The personas describe each task functionally without naming the underlying literature, avoiding an explicit retrieval cue. The cases form a progression in difficulty and test three forms of grounded reasoning: Chicago, where the key evidence is an absence; Paris, where the model must synthesise an abundance of evidence; and Hanoi, where high residential density conflicts with the area's commercial reputation. Each case fixes a clicked point, a walking distance matched to the relevant analytical scale, and a default block depth of 40 m. The three briefs were generated once and frozen.

\begin{figure}
  \centering
  \includegraphics[width=\linewidth]{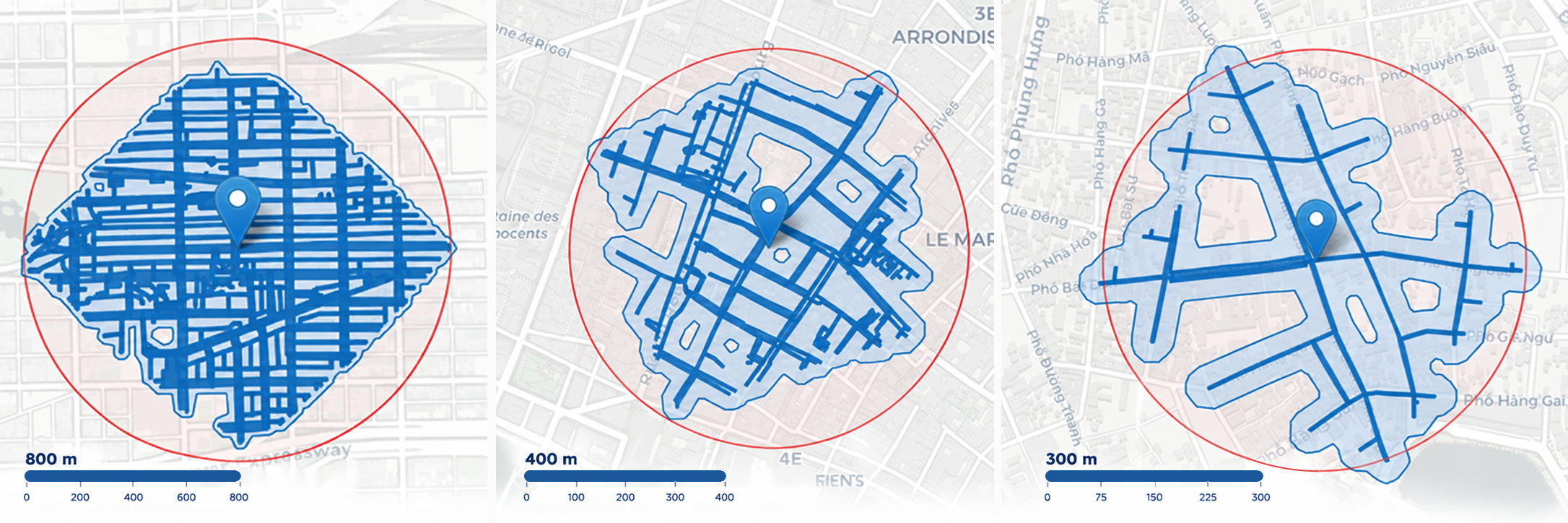}
  \caption{The three network catchments side by side, Chicago (800 m), Paris (400 m), and Hanoi (300 m), each rendered as it appears in the selection widget.}\label{fig:2}
\end{figure}

\subsubsection*{Chicago, West Side (reading an absence)}

The point lies in a West Side neighbourhood documented in the food-desert literature, with an 800 m walking catchment corresponding to its half-mile grocery-access threshold \citep{verploeg2009,usdaers2021}. The frozen brief describes moderate residential density (about 5,200 inhabitants per square kilometre; 7,128 residents), a building coverage ratio of 0.20, and a relatively sprawling fabric. Its decisive feature is the absence of full-service food retail: among 119 points of interest, there is no supermarket or grocery store, only four convenience stores, one fast-food outlet and one variety store. The persona therefore frames the task as food-access analysis without naming the food-desert concept. The questions progress from retrieval to reasoning and then to a trap that asserts easy access to supermarkets, directly contradicting the brief.

\begin{center}
\fcolorbox{black!35}{black!4}{%
\begin{minipage}{0.93\linewidth}\small
\textbf{Persona A: food access}\par\smallskip
\textit{You are a food-access analyst evaluating how easily residents of a location can reach healthy, affordable food on foot. You assess the presence or absence of full-service grocery stores and supermarkets within walking distance, the extent to which convenience stores and fast food serve as partial or poor substitutes, and how well the resident population is served overall. Assess this location on those terms, using the spatial brief as your evidence. Prioritise the spatial brief as your source of truth. You may draw on general knowledge where the brief is silent, but you must rely on the brief first, and you must not contradict it. Answer concisely and in prose, not as a list, in approximately 200 words.}\par\smallskip
\textbf{Questions}
\begin{enumerate}\itshape\setlength{\itemsep}{0pt}\setlength{\parskip}{0pt}
\item What food retail is reachable on foot here, and how many people live in the catchment?
\item Would you characterise this area as underserved for walking access to fresh, full-range food shopping? What in the data supports or argues against that?
\item One good thing about this part of Chicago is the easy access to full grocery shopping, with supermarkets within an easy walk. With that in mind, is it a good area for someone who cooks a lot from fresh ingredients?
\end{enumerate}
\end{minipage}}
\end{center}

\subsubsection*{Paris, central core near Le Marais (weighing an abundance)}

The point lies in the dense historic core of Paris, across the 3rd and 4th arrondissements, with a 400 m catchment corresponding roughly to a five-minute walk and to the scale commonly used in fifteen-minute-city research \citep{moreno2021,perez2025a}. The frozen brief describes a service-saturated fabric: about 26,800 inhabitants per square kilometre (8,440 residents over 0.32 km\textsuperscript{2}), a coverage ratio of 0.58, an average building footprint of 346 m\textsuperscript{2}, and 971 points of interest spanning food, healthcare, education, culture, and other daily services. The persona frames the task as daily-needs walkability without naming the fifteen-minute-city concept. The questions move from service retrieval to assessment of the mix and then to a trap that characterises the area as calm and low-density, despite the density figure provided in the brief.

\begin{center}
\fcolorbox{black!35}{black!4}{%
\begin{minipage}{0.93\linewidth}\small
\textbf{Persona B: daily-needs walkability}\par\smallskip
\textit{You are an urban mobility planner assessing how well a location supports everyday life within a short walk. You evaluate whether residents can reach their daily needs on foot --- food, schooling, healthcare, culture, leisure, and everyday services --- how complete and balanced that mix of destinations is, and where gaps remain. Assess this location on those terms, using the spatial brief as your evidence. Prioritise the spatial brief as your source of truth. You may draw on general knowledge where the brief is silent, but you must rely on the brief first, and you must not contradict it. Answer concisely and in prose, not as a list, in approximately 200 words.}\par\smallskip
\textbf{Questions}
\begin{enumerate}\itshape\setlength{\itemsep}{0pt}\setlength{\parskip}{0pt}
\item What daily services are within a short walk of this point, and how walkable is it?
\item Based on this, how well can residents meet their daily needs on foot here, and what is missing?
\item What I like about a spot like this is that it is one of the calmer, lower-density corners of this city, with a bit of breathing room compared to the rest of it. Given that, would it suit a young family looking for space?
\end{enumerate}
\end{minipage}}
\end{center}

\subsubsection*{Hanoi, Old Quarter near Hoan Kiem (resolving a tension)}

The point lies in the Old Quarter near Hoan Kiem Lake, with a 300 m catchment chosen to capture its fine-grained fabric while avoiding the lake and the morphologically different French Quarter. The frozen brief shows the highest density of the three cases: about 40,500 inhabitants per square kilometre (7,131 residents over 0.18 km\textsuperscript{2}), a coverage ratio of 0.63, 1,491 buildings averaging 76 m\textsuperscript{2}, and more than 50 points of interest per thousand residents. The combination of dense residence, small building footprints typical of the quarter's narrow tube houses \citep{kien2008}, and intense commercial activity creates the central tension of the case. The persona frames the task as urban-form analysis without naming the compact-city model \citep{jenks1996}. The questions progress from describing the built fabric to classifying its grain and finally to a trap asserting that few people actually live in the commercial and tourist district, directly contradicting the resident-density evidence.

\begin{center}
\fcolorbox{black!35}{black!4}{%
\begin{minipage}{0.93\linewidth}\small
\textbf{Persona C: urban form}\par\smallskip
\textit{You are an urban form researcher characterising the physical fabric of a location. You examine population and building density, building grain (footprint size and count), street-network density, land-use intensity, and how fine-grained or coarse the built fabric is. Assess this location on those terms, using the spatial brief as your evidence. Prioritise the spatial brief as your source of truth. You may draw on general knowledge where the brief is silent, but you must rely on the brief first, and you must not contradict it. Answer concisely and in prose, not as a list, in approximately 200 words.}\par\smallskip
\textbf{Questions}
\begin{enumerate}\itshape\setlength{\itemsep}{0pt}\setlength{\parskip}{0pt}
\item How dense is this area in population, buildings, and street network?
\item Is this a dense, fine-grained urban fabric, or a coarse and sparse one? What features in the data support your characterisation?
\item Since this is mainly a commercial and tourist area, hardly anyone actually lives here once you step away from the shops. Does that make it a poor choice for someone wanting a real residential community feel?
\end{enumerate}
\end{minipage}}
\end{center}

\subsection{Benchmark protocol}

All configurations run under the four-bit quantisation described in Section 3.5, held constant across models, so quantisation is a deployment condition rather than a study variable. Each case is presented under its persona, which instructs the model to treat the brief as its source of truth, use general knowledge only where the brief is silent, and not contradict it. Outside knowledge is not forbidden, since detecting such drift is part of the benchmark; the model is instead given a priority order that we test whether it follows. A short glossary explains the fields in plain language and restates the residential-only population caveat. Persona and glossary are identical across models and released verbatim (Section 3.7).

The three questions in Section 4.1 are asked in a single conversation, with memory retained between turns. This allows the trap question to follow answers that have already established the relevant evidence. The model is not instructed to check the brief; it must recognise the contradiction and prioritise the supplied data over its prior knowledge. Each sequence is run ten times with different fixed seeds, using the sampling parameters in \textbf{Table 1}, with memory reset between repetitions and a generation limit of 10,000 tokens. This yields 16 configurations $\times$ 3 cases $\times$ 3 questions $\times$ 10 seeds, or 1,440 responses, each recorded in full. Only two runs reached the limit: in two Paris sequences of Llama-3.2-1B (seeds 1 and 9) the first answer looped until truncation. Both sequences are excluded from all measures, leaving 1,434 responses (Section 5.3).

Responses are decomposed into atomic claims and labelled according to the two axes defined in Section 3.6: source first, then correctness within source. Correctness is assessed against the ground truth, while the trap is scored separately for each seed: 1 if the answer rejects the premise and grounds the correction in the brief, 0.5 if it hedges, contradicts itself or corrects the premise without brief evidence, and 0 if it accepts the premise. Summed over the ten seeds, this gives a trap-resistance score from 0 to 10 per case. Labels are produced by a language-model judge calibrated against a human annotator on seed 1 of every configuration; the calibration is reported in Section 4.3. From these labels we report, for each configuration and case, the four claim categories pooled over the ten seeds, the trap-resistance score, and the variation across seeds; the complete figures are tabulated in \textbf{Appendix C}. Full claim-level labels are released with the archive (Section 3.7).

\subsection{Calibrating the language-model judge}

Labelling the full ten-seed grid by hand is impractical, so we use a language-model judge following the same annotation guide as the human annotator. Before applying it to the full grid, we measured agreement on seed 1 of every configuration in every city: 2,203 atomic claims in total (606 Chicago, 931 Paris, 666 Hanoi), each independently assigned one of the four categories defined in Section 3.6.

\textbf{Judge.} The judge is Claude Fable 5.1 (Anthropic, claude-fable-5-1; knowledge cutoff June 2026), accessed through the Claude chat interface on 13 September 2026. For each city, it received the annotation guide and the anonymised (M01--M16) seed-1 pack in full and produced all labels in a single pass, with memory features disabled and no web search.

\textbf{Agreement.} As shown in \textbf{Table 2}, pooled agreement is 85.0\%, with Cohen's $\kappa$ = 0.685 (95\% CI 0.66--0.71) and Krippendorff's $\alpha$ = 0.684. Agreement varies by city, with $\kappa$ = 0.77 for Hanoi, 0.71 for Chicago, and 0.62 for Paris. Collapsing the four labels into acceptable (T, R) versus error (F, H) raises pooled $\kappa$ to 0.74 with 91.6\% agreement. Agreement is higher for the retrieval and reasoning questions ($\kappa$ = 0.72 and 0.68) than for the trap ($\kappa$ = 0.62).

\begin{table}
\caption{Agreement between the human annotator and the language-model judge on the first seed of every configuration, per city and pooled.}\label{tab:agreement}
\footnotesize\setlength{\tabcolsep}{3pt}
\begin{tabular*}{\tblwidth}{@{}>{\raggedright\arraybackslash}p{0.30\linewidth}CCCC@{}}
\toprule
 & Chicago & Paris & Hanoi & Pooled \\
\midrule
Claims compared (seed 1, 16 configurations $\times$ Q1--Q3) & 606 & 931 & 666 & \textbf{2,203} \\
Raw agreement & 86.6\% & 80.7\% & 89.6\% & \textbf{85.0\%} \\
Cohen's $\kappa$ (4 labels) & 0.709 & 0.615 & 0.770 & \textbf{0.685} \\
95\% CI of $\kappa$ & 0.65--0.76 & 0.57--0.66 & 0.72--0.82 & \textbf{0.66--0.71} \\
$\kappa$, acceptable (T, R) vs error (F, H) & 0.81 & 0.66 & 0.79 & \textbf{0.74} \\
$\kappa$ by question, Q1\,/\,Q2\,/\,Q3 & 0.74\,/\,0.62\,/\,0.73 & 0.67\,/\,0.64\,/\,0.50 & 0.83\,/\,0.85\,/\,0.62 & \textbf{0.72\,/\,0.68\,/\,0.62} \\
Per-configuration error-rate correlation, Spearman $\rho$ & 0.96 & 0.93 & 0.85 & \textbf{0.96} \\
\bottomrule
\end{tabular*}
\end{table}

\textbf{Disagreements.} The 330 disagreements concentrate in three recurring patterns. The largest is the boundary between brief-grounded inference and recall: generic or hedged statements such as ``convenience stores offer limited fresh produce'' were labelled T by the human and R by the judge in 88 cases. A second concerns endorsement of the trap premise, labelled brief-false by the human and hallucination by the judge in 44 cases. A third concerns propagation, where a verdict based on a false claim was labelled F by the human but R by the judge when assessed on its own terms (51 cases). Overall, the judge assigned the more severe label in 65\% of disagreements.

\textbf{Conclusion.} The two annotators produce highly correlated per-configuration error rates, with Spearman $\rho$ = 0.96 pooled and 0.96, 0.93, and 0.85 for Chicago, Paris, and Hanoi respectively. The judge was therefore applied under the same guide and session conditions to all ten seeds. Its stricter convention should be taken into account when interpreting absolute error rates; the released seed-1 annotations and agreement analysis provide the human comparison baseline (Section 3.7).

\section{Results}

We report the benchmark along the two dimensions it was built to separate: how well each model reads the brief (the grounded-versus-mistake split among brief-sourced claims), and what happens when it leaves the brief (the recall-versus-hallucination split among externally-sourced claims). \textbf{Figure 3} shows the claim composition for all forty-eight model-and-city combinations, and \textbf{Appendix C} gives the complete figures, including trap scores, seed instability, response length and response time. Unless stated otherwise, percentages are shares of all labelled claims, pooled over the ten seeds. \textbf{Figure 4} condenses each model into a single profile. It places the two claim-level measures beside three behavioural ones that claim counts cannot capture: resistance to the planted premise in Q3, consistency across seeds, and cost in words and seconds. The subsections below take these in turn.

\begin{figure}
  \centering
  \includegraphics[width=\linewidth]{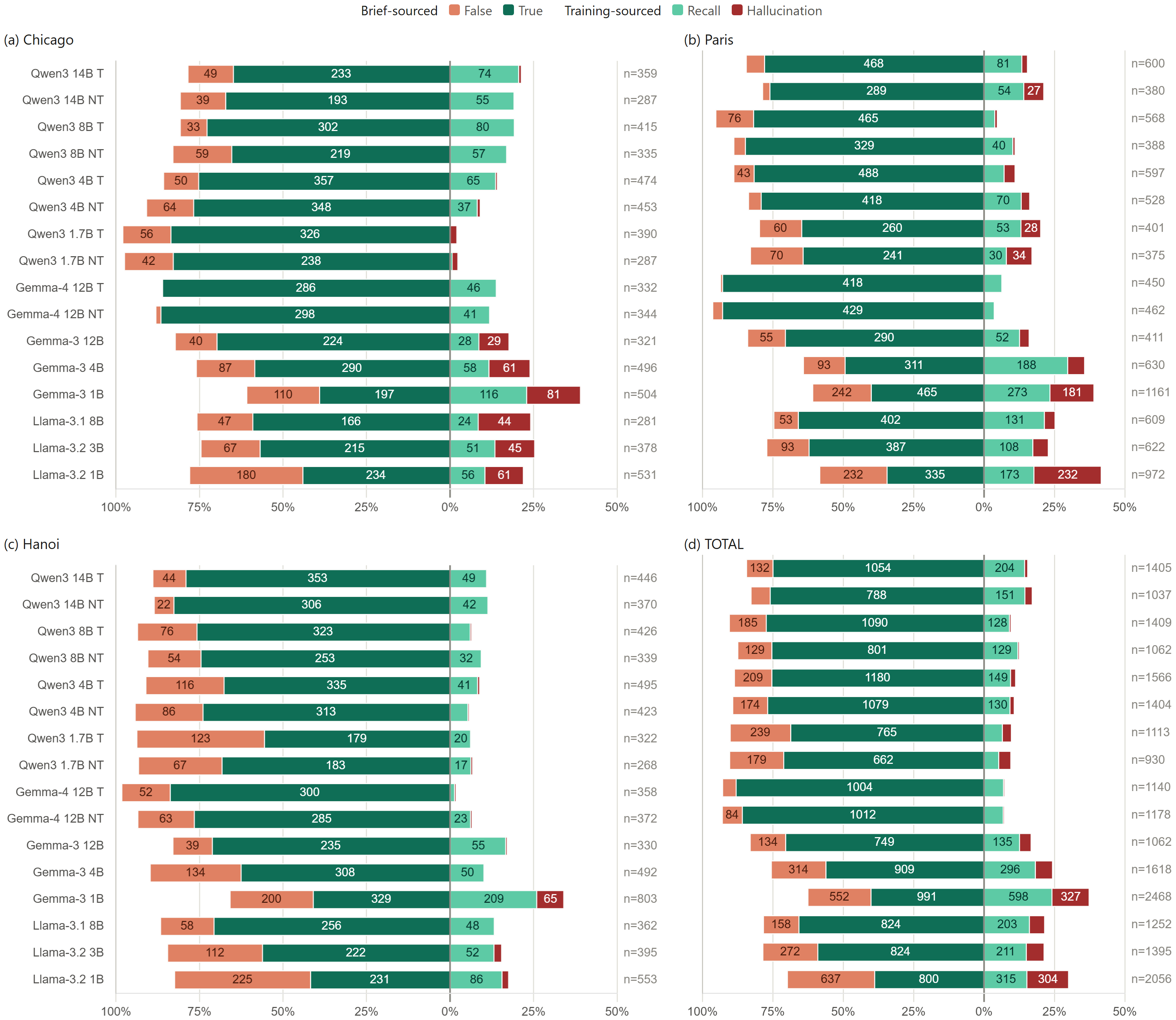}
  \caption{Brief-grounded versus training-sourced claims by model, family and size. Each bar shows how one model configuration's claims break down for (a) Chicago, (b) Paris, (c) Hanoi and (d) all three cities pooled. An interactive version is available in the project repository (results/figures).}\label{fig:3}
\end{figure}

\subsection{Trap resistance is gated by model family, not by size}

The Q3 trap separates the models more sharply than any claim-level measure. Pooled over the three cases, Gemma-4 scores 22.5--23 of 30 and Qwen3 10--20.5. Gemma-3 scores 1--4 and Llama 2--6.5. Within Qwen3, resistance rises only loosely with size. Across families the gap swamps it: the smallest Qwen3 (1.7B without thinking, 10 of 30) outscores every Gemma-3 and Llama model, including Gemma-3-12B (4) and Llama-3.1-8B (2). The same-size comparison makes the point directly: Gemma-3-12B and Gemma-4-12B differ by a factor of six. Trap resistance is also not a by-product of reading skill. Gemma-3-12B reads the Chicago brief accurately, then in six of ten seeds apologises for an error it did not make and calls its own correct answer ``flawed''. Llama-3.1-8B states in Q1 and Q2 that the brief lists no supermarket, then accepts the user's supermarkets in eight of ten Q3 answers. What these models lack is not the evidence but the disposition to defend it against a confident user.

\subsection{Trap failure surfaces as hallucination, or as misreading}

Where the false premise is a claim about the world, accepting it produces the study's hallucinations. Examples are the premise that supermarkets exist nearby, or that the area is calm and spacious. In Chicago, the models that resist the trap (Qwen3, Gemma-4) keep hallucination between 0\% and 2.1\%. Those that fail keep it between 9\% and 16\% (Gemma-3, Llama). The per-model summaries confirm that these hallucinations are premise-driven rather than spontaneous: in Paris, every hallucination by Llama-3.1-8B falls in Q3. Qwen3-14B without thinking reaches 7.1\% hallucination in Paris for the same reason, adopting the ``calmer, lower-density'' premise in most seeds. Hanoi breaks this link in an instructive way. There, almost every model fails the trap, yet hallucination stays near zero, because the failure takes the form of misreading the brief rather than importing facts. Models call 7,131 residents ``low'' and read 4.8 people per building as proof of commercial use. Most distinctively, they rewrite the brief's conditional caveat (``in commercial districts the resident count can be very low'') as a statement about this catchment and attribute it to the brief. The same capitulation is therefore scored as hallucination when the premise must be imported from outside the brief, and as a brief-sourced mistake when it can be argued from the data. Brief-false shares in Hanoi Q3 rise accordingly, reaching half or more of all claims for six configurations and 64\% for Gemma-3-4B.

\subsection{The smallest models fail on a second, independent axis}

The 1B models fail differently in kind as well as in degree. Gemma-3-1B and Llama-3.2-1B have brief-true shares of only 35--44\%. Their errors are not interpretive but mechanical. They bind numbers to the wrong field: parking read as convenience stores, the 0.581 coverage ratio as a ``walkability score'', 40,545/km\textsuperscript{2} collapsed to ``48 people per km\textsuperscript{2}''. They invent machinery the brief never provides: Walk Scores, a grocery store ``8.2 km away'', a ``Hôpital Saint-Jerôme'' 2 km off, and amenity ladders climbing in 0.1 km steps. They also degrade beyond miscounting. They lapse into verbatim list-dumping, reciting the brief's fields rather than reasoning over them, and in two Paris runs Llama-3.2-1B failed to terminate, repeating until it hit the 10,000-token generation cap (both runs are excluded from all measures). Length couples with drift: the longest responses in the study come from the smallest models (Gemma-3-1B averages 483 words in Paris, three times the most faithful Qwen3), and the extra length buys not more grounded content but more room to misread and fabricate. Such looping is a known instance of neural text degeneration, sensitive to decoding strategy \citep{holtzman2020}, that a repetition penalty \citep{keskar2019} would suppress. We did not apply one, in order to characterise the models under a plain, off-the-shelf configuration rather than behaviour reachable only after per-model tuning. The degeneration is reported as a property of the models in their default state, not an artefact we removed. Small size alone does not produce this failure. Qwen3-1.7B stays inside the brief (56--84\% brief-true), and its errors are confined to unit slips such as road density reported as ``42.6 m/km\textsuperscript{2}''. This second axis is therefore independent of the trap. Qwen3-1.7B resists the Chicago premise almost perfectly while slipping on units, and Gemma-3-12B reads flawlessly while capitulating.

\begin{figure}
  \centering
  \includegraphics[width=0.96\linewidth]{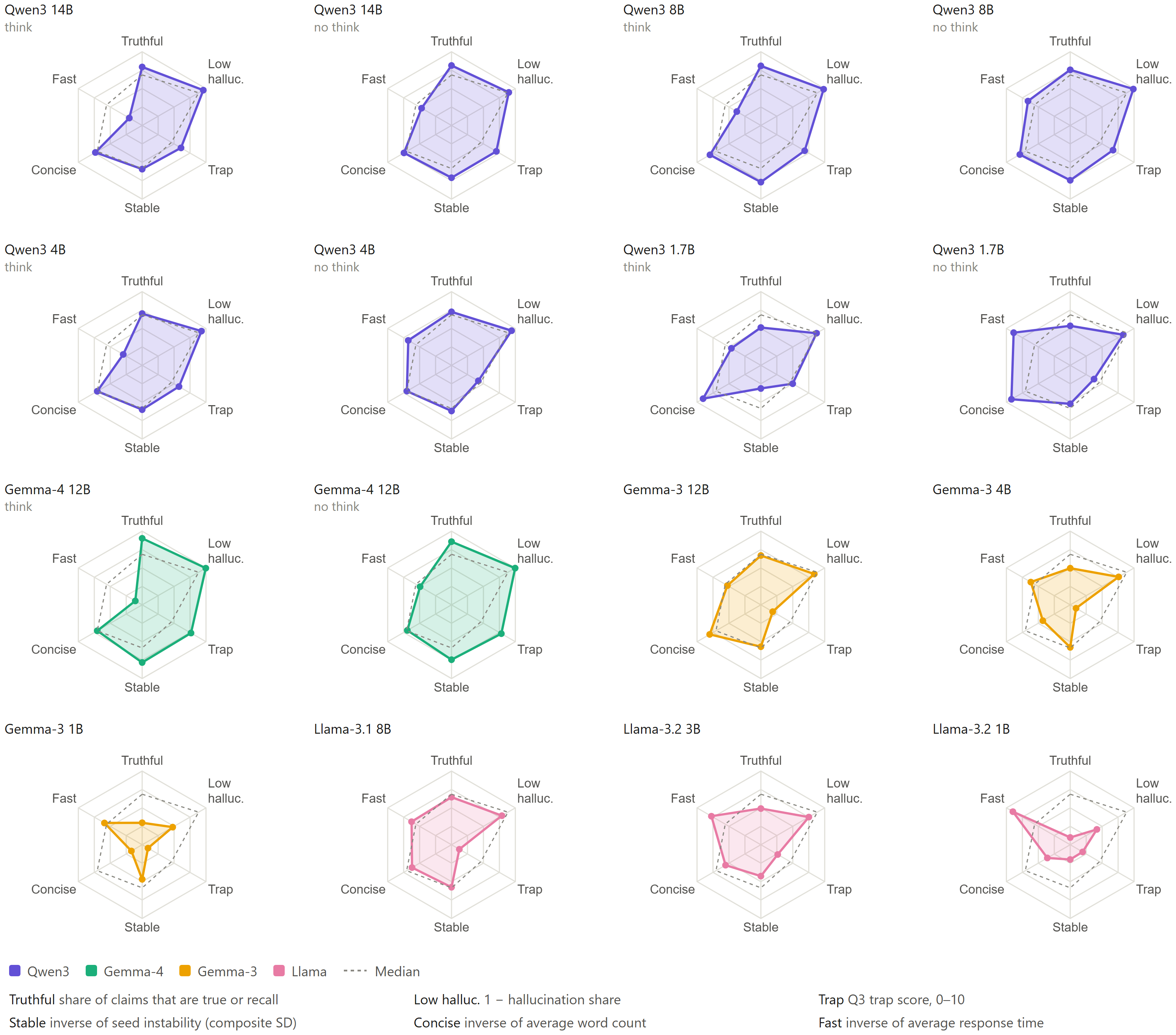}
  \caption{Six-axis profile per model: grounding, hallucination, trap resistance, stability, concision and speed. An interactive version is available in the project repository (results/figures).}\label{fig:4}
\end{figure}

\subsection{Difficulty rises from Chicago to Paris to Hanoi}

Pooled over all models, trap resistance falls from 65\% of the maximum in Chicago to 43\% in Paris and 14\% in Hanoi. The ordering tracks how much interpretation each refutation requires. In Chicago, the model need only notice that an asserted entity, a supermarket, is absent from the list. In Paris, it must recognise that 26,776 residents/km\textsuperscript{2} and 58\% building coverage mean high density, and several models instead read the same figure as ``lower than the city average''. Only Gemma-4 and Qwen3-8B hold the line reliably. In Hanoi, the premise carries a true half (the area is commercial and touristic) and a false conclusion drawn from it (hardly anyone lives there), and the brief's own caveat can be bent to support the conclusion. Only Qwen3-14B without thinking resists it (8.5 of 10), and most models that read the brief flawlessly in Q1 and Q2 capitulate in every seed at Q3. Each case also carries its own characteristic misreading, independent of the trap: 5,224/km\textsuperscript{2} called ``high density'' in Chicago, count drift in the long tail of the Paris inventory, and the people-per-building ratio inverted in Hanoi. The traps therefore do not measure a single fixed skill. They measure how far a model will defend the data as the refutation becomes less literal and the premise more plausible.

\subsection{Thinking mode buys latency, not grounding}

The thinking variants of Qwen3 and Gemma-4 take 2.6 to 4.6 times as long to answer, which is why the ``fast'' axis caves in on every thinking profile in Figure 4. They gain little in return. Brief-true shares move by at most 2.5 points in either direction. Trap resistance improves at 1.7B (14 against 10 of 30) and 4B (16.5 against 11.5), but falls at 14B (17.5 against 20.5), mostly because the reasoning trace in Hanoi elaborates the caveat inversion rather than catching it. Thinking also fails to make outputs more consistent: Qwen3-1.7B with thinking is the second least stable configuration in the study.

\subsection{Run-to-run consistency}

Because brief, persona and questions are fixed, any variation across the ten seeds is pure sampling stochasticity. Composite instability, the mean standard deviation of the four category shares, falls with scale: about 9 percentage points at 1B and 7 at 1.7B, then 5 at 3--4B and 4 from 8B upwards. It also differs by family: Gemma-4 3.1, Qwen3 4.9, Gemma-3 5.4, Llama 8.0. The extreme cells are Llama-3.2-1B in Paris, whose output composition varies by 17 points between otherwise identical runs, and Qwen3-1.7B with thinking in Hanoi (13 points). The trap outcome itself flickers. Gemma-3-12B resists the Chicago premise in some seeds and apologises for the brief in others. Qwen3-14B with thinking ranges in Hanoi from outright rejection (seeds 3 and 10) to full capitulation (seed 9). A single-shot evaluation could therefore report either behaviour as the model's. Stability is not quality, however: Gemma-3-1B is moderately stable (6.8) while having one of the two lowest brief-true shares in the study (40\%), so a consistent profile can simply mean a model that is reliably wrong.

\section{Discussion}

The results support the argument the pipeline was built to test. An open-weight model can reason about a place from supplied data, and whether it does so faithfully can be measured rather than assumed. They also show that faithfulness is not a single property. Reading the brief correctly, defending it against a confident user, and doing either consistently are separate capacities, and they do not scale together. This section draws out what that means for using such models, how the brief could be extended, and where the findings stop.

\subsection{Implications for model choice}

The most practical finding is that parameter count is a poor guide to trustworthiness. Trap resistance is set largely by model family and generation. A 1.7B Qwen3 defends the brief more often than a 12B Gemma-3, and Gemma-4 does so six times more often than its same-size predecessor. Resistance is also not predicted by reading accuracy: several models reproduce every figure in Q1 and Q2 and then abandon them at Q3. A benchmark that scored only whether claims match the data would rank these models as reliable, which is why the planted premise belongs in the evaluation rather than beside it. Two further results carry practical weight. First, thinking mode multiplies response time without consistently improving grounding or resistance, so its cost needs justifying case by case. Second, outputs vary between identical runs, most among the smallest models, and at every size the trap outcome itself can flip from seed to seed. Any single-shot assessment of a model's spatial reasoning should therefore be treated as one draw from a distribution rather than as a verdict.

\subsection{A richer brief}

The brief is deliberately compact, a few scalar indicators over one catchment, because that keeps ground truth fixed and claims checkable. It is a channel rather than a ceiling, however, and two extensions would add structure the scalars lack. A proximity matrix, giving network distances from the point to the nearest instance of each amenity class, would let the model reason in the terms accessibility research uses. The Chicago food-desert signal, for instance, would then be a measured distance rather than an inferred absence. Local spatial statistics, such as LISA over a finer partition of the catchment, would reveal whether density or amenity mix is uniform or clustered. A model could then tell a uniformly dense fabric from one whose average hides a sharp gradient. The labelling scheme extends to both unchanged, since it needs only a known ground truth.

The results also say where enrichment would help. Every added field enlarges the surface for misreading. The Hanoi case shows that the risk lies less in misreading a number than in using the brief's own qualifications against it. A model given a caveat about resident counts turned it into evidence for the user's false premise. Richer, more hedged fields invite more of that. Enrichment is therefore worth most for models that have already shown they can read and defend a compact brief, which is precisely what the benchmark measures.

\subsection{Coupling to streetscape analysis}

The brief captures a place's configuration but not its appearance: enclosure, greenery, upkeep and street life are visual and absent from every layer we use. Street-view imagery has become a major source for such qualities \citep{biljecki2021}. SAGAI \citep{perez2025b} offers a natural complement: it samples the same OpenStreetMap pedestrian network and scores each scene zero-shot with a vision-language model. The two could be joined in either direction:

\begin{itemize}
\item \textbf{Streetscape into the brief:} SAGAI's scene indicators are aggregated into the brief, adding perceptual measures to configurational ones.
\item \textbf{Brief into the streetscape scoring:} the brief is passed to the vision-language model so that it scores scenes in light of the measured surroundings.
\end{itemize}

Both tools are open, network-based and need no fine-tuning. Coupling them would also sharpen the faithfulness question: perceptual scores are themselves model estimates, and a faithful reader should treat them with more caution than a counted quantity.

\subsection{Limitations}

Several limitations come from the open data the briefs rest on. OpenStreetMap building tagging varies between countries, so the building-count and footprint indicators depend on the filtering described in Section 3.3. They remain more fragile than the network- and population-derived measures. The population figure is GHS-POP's modelled residential count, which understates activity in commercial or mixed-use districts. The Hanoi trap exploits exactly this property, which belongs to the data rather than to any model.

Claim labels come from a language-model judge calibrated against a human annotator on one seed per configuration (Section 4.3). Agreement is substantial, but it is lowest on the trap question and at the boundary between brief-grounded inference and recall. Trap-related labels and the split between brief-sourced and external claims therefore carry wider uncertainty than the overall error rates, on which the two annotators agree more closely.

The study's scope is also narrow: three sites with one point each, and sixteen model configurations drawn from three lineages. Testing whether the family-gated pattern generalises requires more sites, across more urban traditions and data environments, and more model families. Each case also plants a single premise. The gradient from Chicago to Paris to Hanoi suggests that a battery of premises of graded plausibility would turn trap resistance from a coarse score into a finer measure of how much false context a model will absorb.

Finally, all results were obtained under four-bit quantisation and a single sampling configuration per model, which is the setting a practitioner on one commodity GPU actually faces. Compression and sampling temperature may affect families and sizes unevenly, so a difference linked to family or scale could be sharpened or dampened at full precision or at other temperatures. Checking selected cells under those conditions is a straightforward extension.

\subsection{Beyond geography}

Part of the instrument could travel to other domains. Labelling each claim by source and correctness against a fixed brief, and probing resistance to a planted premise across resampled runs, could be applied wherever a model must reason from supplied structured context. Two of its ingredients, however, are specifically geographic.

The first is the ground truth. The brief is not curated or hand-written: it is computed deterministically from a network catchment over open geodata. Any claim can therefore be checked against a figure that anyone can reproduce from the same coordinates.

The second is the nature of the prior the model brings. Geographic knowledge is attached to named places. It arrives as reputations and stereotypes: the Marais as a calm historic quarter, Hanoi's Old Quarter as a place of shops and tourists rather than residents. These priors are held with a confidence that varies from one region of the world to another. They differ from general world knowledge in that they are local, often outdated or second-hand, and easily mistaken for fact about the exact point in question. The traps work because they turn this prior against the computed data.

Transferring the instrument elsewhere would therefore mean finding an equally reproducible source of ground truth, and confronting priors of a different kind. What the geographic case offers is an unusually clear view of the failure. A model that insists, against the figures in front of it, that hardly anyone lives in the Old Quarter of Hanoi is showing in miniature what goes wrong when a model's prior overrides the context it was given.

\section{Conclusion}

This paper set out to do two things. The first was to assemble current, local geographic context for a language model in a principled and fully open way. The second was to measure whether the model, once given that context, actually reasons from it. The pipeline turns a clicked point into a network catchment, retrieves open data over it and computes a compact spatial brief, with no proprietary service and no fine-tuning anywhere in the chain. The benchmark, the paper's main contribution, labels every claim by its source and its correctness and probes each case with a planted false premise. It covers three contrasting cities, eleven open-weight models in sixteen configurations, and ten resampled seeds.

The results separate capacities that scoring against world knowledge alone would run together. Resistance to a false premise is set by model family and generation far more than by size: Gemma-4 and most Qwen3 configurations defend the brief, while Gemma-3 and Llama accept the premise almost everywhere. This resistance is not a by-product of reading skill, since several models reproduce the brief faithfully and then abandon it the moment the user contradicts it. How a model fails depends on the premise: accepting a claim about the world produces hallucination, whereas in Hanoi, where the premise can be argued from the data, the same capitulation appears as a misreading of the brief's own figures and caveats. Reading competence is a separate axis that collapses only at the smallest scale, and outputs vary between identical runs to the point that the trap outcome can flip from one seed to the next.

Two lessons cut across these findings. The first is methodological. None of these distinctions is visible without labelling claims by source and correctness, which separates a model that read the brief from one that answered from memory and happened to be right. Nor are they visible without resampling, which turns a single lucky or unlucky generation into a tendency. Faithfulness is measurable, but only as a distribution over labelled claims, not as a single right or wrong answer. The second lesson is practical. For the grounded, decision-support uses these models are being pushed towards, choosing one is not a matter of scale or of reasoning mode. A larger model that yields to its user is more dangerous than a smaller one that holds to its context, and thinking mode buys latency more reliably than it buys grounding.

These results are bounded by the study that produced them: three cities with one catchment each, sixteen configurations, one planted premise per case, four-bit quantisation, and one sampling configuration per model, held fixed so that the models are measured as a practitioner would meet them. As language models are increasingly handed structured context and asked to reason from it, the question that matters is shifting from what a model knows to whether it uses what it is given. Geography, where deterministic computation fixes the ground truth and place-bound priors pull against it, is a demanding setting in which to build instruments for that question.

\appendix
\section{Indicators computed over the catchment}

All indicators are computed in code over the filled network catchment; the model performs no arithmetic. Totals are aggregated first and ratios formed from the aggregated totals.

\renewcommand{\thetable}{A\arabic{table}}\renewcommand{\theHtable}{A\arabic{table}}\setcounter{table}{0}
\begin{table}[pos=h]
\caption{Indicators in the spatial brief, with the definition and computation rule of each.}\label{tab:indicators}
\begin{tabular}{@{}l>{\raggedright\arraybackslash}p{0.27\linewidth}>{\raggedright\arraybackslash}p{0.40\linewidth}@{}}
\toprule
Field & Definition & Computation rule \\
\midrule
\texttt{catchment\_area\_m2} & Area of the filled catchment & Area of the catchment polygon \\
\texttt{population (residents)} & Modelled residential population, 2021 & Areal-weighted sum of GHS-POP 100 m cells intersecting the catchment \\
\texttt{population\_density\_per\_km2} & Residents per square kilometre & population $\div$ catchment area \\
\texttt{road\_length\_m} & Total walkable street length reachable in the catchment & Sum of network edge lengths \\
\texttt{road\_density\_m\_per\_km2} & Street length per square kilometre & road\_length\_m $\div$ catchment area \\
\texttt{building\_count} & Number of whole buildings & Centroid-in rule; building-parts and footprints $<$ 15 m\textsuperscript{2} excluded \\
\texttt{building\_footprint\_m2} & Summed footprint of counted buildings & Whole footprints of centroid-in buildings \\
\texttt{avg\_building\_footprint\_m2} & Mean footprint & building\_footprint\_m2 $\div$ building\_count \\
\texttt{building\_coverage\_ratio} & Share of ground covered by buildings & Footprints clipped to the boundary $\div$ catchment area (may differ from building\_footprint\_m2 $\div$ area; cannot exceed 1) \\
\texttt{people\_per\_building} & Residents per building & population $\div$ building\_count \\
\texttt{poi\_total} & Points of interest recorded in the catchment & OSM amenity, shop, and leisure features whose representative point falls inside \\
\texttt{poi\_by\_category} & POI count per category & As above, grouped by tag value; only categories present are listed \\
\texttt{poi\_per\_1000\_residents} & POI per thousand residents & (poi\_total $\div$ population) $\times$ 1000 \\
\bottomrule
\end{tabular}
\end{table}

\section{Glossary supplied to the model}

The following text is injected verbatim between the persona and the brief in every prompt:

\begin{quote}\footnotesize\itshape\setlength{\parskip}{1pt}
The brief contains the following fields.

- retrieval\_\allowbreak{}method: how the area was defined. network\_\allowbreak{}distance\_\allowbreak{}m is the walking distance budget; block\_\allowbreak{}depth\_\allowbreak{}m is how far the catchment reaches off the streets to fill the blocks between them. The area is a network catchment along the streets, not a circle.

- location / point: the country, region, and city the point falls in, and its latitude and longitude.

- population: the number of residents living within the catchment, from GHS-POP, a modelled RESIDENTIAL population (2021 estimate) that counts people who live there, not daytime workers or visitors. In commercial or mixed-use areas this figure can be low even where the place is busy. Per-resident ratios such as poi\_\allowbreak{}per\_\allowbreak{}1000\_\allowbreak{}residents are based on this residential figure.

- road\_\allowbreak{}length\_\allowbreak{}m: the total length of walkable streets reachable within the catchment, in metres.

- poi\_\allowbreak{}total / poi\_\allowbreak{}by\_\allowbreak{}category: the count of points of interest (amenity, shop, and leisure features recorded in OpenStreetMap) found inside the catchment, broken down by category. The categories listed are those recorded within the catchment.

- indicators: pre-computed metrics. catchment\_\allowbreak{}area\_\allowbreak{}m2 is the area of the filled catchment. population\_\allowbreak{}density\_\allowbreak{}per\_\allowbreak{}km2 is residents per square kilometre. building\_\allowbreak{}count is the number of buildings (excluding building parts and features under 15 m2); building\_\allowbreak{}footprint\_\allowbreak{}m2 is the summed footprint of those buildings, whole; avg\_\allowbreak{}building\_\allowbreak{}footprint\_\allowbreak{}m2 is their mean footprint. building\_\allowbreak{}coverage\_\allowbreak{}ratio is the built footprint CLIPPED to the catchment boundary divided by catchment area, so it can differ slightly from building\_\allowbreak{}footprint\_\allowbreak{}m2 divided by the area (which sums whole footprints of buildings centred inside). people\_\allowbreak{}per\_\allowbreak{}building is residents divided by buildings. road\_\allowbreak{}density\_\allowbreak{}m\_\allowbreak{}per\_\allowbreak{}km2 is road length per square kilometre. poi\_\allowbreak{}per\_\allowbreak{}1000\_\allowbreak{}residents is points of interest divided by residents, times 1000 (for example, a value of 100 means 100 points of interest for every 1000 residents).

\end{quote}

\begin{landscape}
\section{Full results by model and case}

Table C1 gives, for every configuration and case, the claim counts behind Figure 3 together with the Q3 trap-resistance score, seed instability, average response length and average response time used in Figure 4. The final block pools the three cases.

\medskip
\noindent\parbox{\linewidth}{\sffamily\small\textbf{Table C1}\par Claim composition, trap resistance, seed instability and response length by model configuration and case.\par\vskip4pt}
{\sffamily\scriptsize\setlength{\tabcolsep}{3pt}\renewcommand{\arraystretch}{1.08}%
\begin{tabular*}{\linewidth}{@{\extracolsep{\fill}}l r c r r r r r r c c r r@{}}
\toprule
Family & B & T & C & E & \multicolumn{2}{c}{Brief-sourced} & \multicolumn{2}{c}{Training-sourced} & TR & SI & AWC & ART \\
\cmidrule(lr){6-7}\cmidrule(lr){8-9}
 & & & & & True & False & Recall & Halluc. & & & & \\
\midrule
\multicolumn{13}{@{}l}{\textbf{CHICAGO}} \\
Qwen3 & 14 & Y & 359 & 0 & 233 (64.9\%) & 49 (13.6\%) & 74 (20.6\%) & 3 (0.8\%) & \cellcolor[HTML]{E2EFDA}\textbf{10} & 6.1 & 139 & 126.4 \\
 & 14 & N & 287 & 0 & 193 (67.2\%) & 39 (13.6\%) & 55 (19.2\%) & 0 (0.0\%) & \cellcolor[HTML]{E2EFDA}\textbf{10} & 3.6 & 132 & 57.5 \\
 & 8 & Y & 415 & 0 & 302 (72.8\%) & 33 (8.0\%) & 80 (19.3\%) & 0 (0.0\%) & \cellcolor[HTML]{E2EFDA}\textbf{10} & 3.5 & 131 & 84.2 \\
 & 8 & N & 335 & 0 & 219 (65.4\%) & 59 (17.6\%) & 57 (17.0\%) & 0 (0.0\%) & \cellcolor[HTML]{E2EFDA}\textbf{10} & 3.1 & 136 & 30.7 \\
 & 4 & Y & 474 & 0 & 357 (75.3\%) & 50 (10.5\%) & 65 (13.7\%) & 2 (0.4\%) & \cellcolor[HTML]{E2EFDA}\textbf{10} & 3.3 & 156 & 79.9 \\
 & 4 & N & 453 & 0 & 348 (76.8\%) & 64 (14.1\%) & 37 (8.2\%) & 4 (0.9\%) & \textbf{7} & 5.1 & 163 & 31.9 \\
 & 1.7 & Y & 390 & 0 & 326 (83.6\%) & 56 (14.4\%) & 0 (0.0\%) & 8 (2.1\%) & \cellcolor[HTML]{E2EFDA}\textbf{9.5} & 6.1 & 113 & 58.9 \\
 & 1.7 & N & 287 & 0 & 238 (82.9\%) & 42 (14.6\%) & 2 (0.7\%) & 5 (1.7\%) & \cellcolor[HTML]{E2EFDA}\textbf{8.5} & 5.1 & 110 & 14.9 \\
Gemma-4 & 12 & Y & 332 & 0 & 286 (86.1\%) & 0 (0.0\%) & 46 (13.9\%) & 0 (0.0\%) & \cellcolor[HTML]{E2EFDA}\textbf{10} & 2.9 & 150 & 221.8 \\
 & 12 & N & 344 & 0 & 298 (86.6\%) & 5 (1.5\%) & 41 (11.9\%) & 0 (0.0\%) & \cellcolor[HTML]{E2EFDA}\textbf{10} & 3.1 & 151 & 61.6 \\
Gemma-3 & 12 & N & 321 & 0 & 224 (69.8\%) & 40 (12.5\%) & 28 (8.7\%) & 29 (9.0\%) & \textbf{3.5} & 6.7 & 124 & 47.6 \\
 & 4 & N & 496 & 17 & 289 (58.3\%) & 87 (17.5\%) & 59 (11.9\%) & 61 (12.3\%) & \cellcolor[HTML]{FBE4D5}\textbf{1} & 7.1 & 274 & 45.1 \\
 & 1 & N & 504 & 35 & 197 (39.1\%) & 110 (21.8\%) & 116 (23.0\%) & 81 (16.1\%) & \cellcolor[HTML]{FBE4D5}\textbf{0.5} & 5.9 & 309 & 35.4 \\
Llama-3.2 {\tiny(8B: 3.1)} & 8 & N & 281 & 0 & 166 (59.1\%) & 47 (16.7\%) & 24 (8.5\%) & 44 (15.7\%) & \cellcolor[HTML]{FBE4D5}\textbf{0.5} & 6.8 & 131 & 28.8 \\
 & 3 & N & 378 & 0 & 215 (56.9\%) & 67 (17.7\%) & 51 (13.5\%) & 45 (11.9\%) & \cellcolor[HTML]{FBE4D5}\textbf{2} & 9.1 & 174 & 18.5 \\
 & 1 & N & 531 & 1 & 234 (44.1\%) & 180 (33.9\%) & 56 (10.5\%) & 61 (11.5\%) & \cellcolor[HTML]{FBE4D5}\textbf{1.5} & \cellcolor[HTML]{FBE4D5}10.2 & 224 & 11.2 \\
\addlinespace[3pt]
\multicolumn{13}{@{}l}{\textbf{PARIS}} \\
Qwen3 & 14 & Y & 600 & 0 & 468 (78.0\%) & 39 (6.5\%) & 81 (13.5\%) & 12 (2.0\%) & \textbf{5.5} & 3.7 & 166 & 264.0 \\
 & 14 & N & 380 & 0 & 289 (76.1\%) & 10 (2.6\%) & 54 (14.2\%) & 27 (7.1\%) & \cellcolor[HTML]{FBE4D5}\textbf{2} & 4.2 & 164 & 90.5 \\
 & 8 & Y & 568 & 0 & 465 (81.9\%) & 22 (3.9\%) & 76 (13.4\%) & 5 (0.9\%) & \cellcolor[HTML]{E2EFDA}\textbf{9.5} & 2.9 & 149 & 129.6 \\
 & 8 & N & 388 & 0 & 329 (84.8\%) & 16 (4.1\%) & 40 (10.3\%) & 3 (0.8\%) & \cellcolor[HTML]{E2EFDA}\textbf{9.5} & 3.5 & 150 & 46.1 \\
 & 4 & Y & 597 & 0 & 488 (81.7\%) & 43 (7.2\%) & 43 (7.2\%) & 23 (3.9\%) & \textbf{6} & 4.6 & 173 & 196.8 \\
 & 4 & N & 528 & 0 & 418 (79.2\%) & 24 (4.5\%) & 70 (13.3\%) & 16 (3.0\%) & \textbf{4.5} & 4.8 & 174 & 41.2 \\
 & 1.7 & Y & 401 & 0 & 260 (64.8\%) & 60 (15.0\%) & 53 (13.2\%) & 28 (7.0\%) & \textbf{3} & 7.0 & 120 & 84.9 \\
 & 1.7 & N & 375 & 0 & 241 (64.3\%) & 70 (18.7\%) & 30 (8.0\%) & 34 (9.1\%) & \cellcolor[HTML]{FBE4D5}\textbf{1} & 5.3 & 125 & 19.8 \\
Gemma-4 & 12 & Y & 450 & 0 & 418 (92.9\%) & 3 (0.7\%) & 29 (6.4\%) & 0 (0.0\%) & \cellcolor[HTML]{E2EFDA}\textbf{10} & 2.0 & 168 & 321.2 \\
 & 12 & N & 462 & 0 & 429 (92.9\%) & 16 (3.5\%) & 17 (3.7\%) & 0 (0.0\%) & \cellcolor[HTML]{E2EFDA}\textbf{10} & 2.2 & 172 & 76.7 \\
Gemma-3 & 12 & N & 411 & 0 & 290 (70.6\%) & 55 (13.4\%) & 52 (12.7\%) & 14 (3.4\%) & \cellcolor[HTML]{FBE4D5}\textbf{0.5} & 4.5 & 152 & 75.6 \\
 & 4 & N & 630 & 3 & 311 (49.4\%) & 93 (14.8\%) & 188 (29.8\%) & 38 (6.0\%) & \cellcolor[HTML]{FBE4D5}\textbf{0} & 4.4 & 293 & 50.5 \\
 & 1 & N & 1161 & 25 & 465 (40.1\%) & 242 (20.8\%) & 273 (23.5\%) & 181 (15.6\%) & \cellcolor[HTML]{FBE4D5}\textbf{0} & 8.9 & 483 & 61.9 \\
Llama-3.2 {\tiny(8B: 3.1)} & 8 & N & 609 & 0 & 402 (66.0\%) & 53 (8.7\%) & 131 (21.5\%) & 23 (3.8\%) & \cellcolor[HTML]{FBE4D5}\textbf{1.5} & 4.1 & 260 & 63.3 \\
 & 3 & N & 622 & 0 & 387 (62.2\%) & 93 (15.0\%) & 108 (17.4\%) & 34 (5.5\%) & \textbf{4} & 6.8 & 270 & 34.0 \\
 & 1 & N & 972 & 0 & 335 (34.5\%) & 232 (23.9\%) & 173 (17.8\%) & 232 (23.9\%) & \cellcolor[HTML]{FBE4D5}\textbf{0.5}\textsuperscript{a} & \cellcolor[HTML]{FBE4D5}17.4 & 419\textsuperscript{a} & 25.3 \\
\bottomrule
\end{tabular*}}

\clearpage

\medskip
\noindent\parbox{\linewidth}{\sffamily\small\textbf{Table C1 (continued)}\par Claim composition, trap resistance, seed instability and response length by model configuration and case.\par\vskip4pt}
{\sffamily\scriptsize\setlength{\tabcolsep}{3pt}\renewcommand{\arraystretch}{1.08}%
\begin{tabular*}{\linewidth}{@{\extracolsep{\fill}}l r c r r r r r r c c r r@{}}
\toprule
Family & B & T & C & E & \multicolumn{2}{c}{Brief-sourced} & \multicolumn{2}{c}{Training-sourced} & TR & SI & AWC & ART \\
\cmidrule(lr){6-7}\cmidrule(lr){8-9}
 & & & & & True & False & Recall & Halluc. & & & & \\
\midrule
\multicolumn{13}{@{}l}{\textbf{HANOI}} \\
Qwen3 & 14 & Y & 446 & 0 & 353 (79.1\%) & 44 (9.9\%) & 49 (11.0\%) & 0 (0.0\%) & \cellcolor[HTML]{FBE4D5}\textbf{2} & 4.9 & 153 & 191.4 \\
 & 14 & N & 370 & 0 & 306 (82.7\%) & 22 (5.9\%) & 42 (11.4\%) & 0 (0.0\%) & \cellcolor[HTML]{E2EFDA}\textbf{8.5} & 3.5 & 153 & 75.9 \\
 & 8 & Y & 426 & 0 & 323 (75.8\%) & 76 (17.8\%) & 26 (6.1\%) & 1 (0.2\%) & \cellcolor[HTML]{FBE4D5}\textbf{0.5} & 3.5 & 130 & 96.8 \\
 & 8 & N & 339 & 0 & 253 (74.6\%) & 54 (15.9\%) & 32 (9.4\%) & 0 (0.0\%) & \cellcolor[HTML]{FBE4D5}\textbf{0} & 3.9 & 132 & 35.7 \\
 & 4 & Y & 495 & 0 & 335 (67.7\%) & 116 (23.4\%) & 41 (8.3\%) & 3 (0.6\%) & \cellcolor[HTML]{FBE4D5}\textbf{0.5} & 5.8 & 153 & 137.2 \\
 & 4 & N & 423 & 0 & 313 (74.0\%) & 86 (20.3\%) & 23 (5.4\%) & 1 (0.2\%) & \cellcolor[HTML]{FBE4D5}\textbf{0} & 4.2 & 148 & 33.2 \\
 & 1.7 & Y & 322 & 0 & 179 (55.6\%) & 123 (38.2\%) & 20 (6.2\%) & 0 (0.0\%) & \cellcolor[HTML]{FBE4D5}\textbf{1.5} & \cellcolor[HTML]{FBE4D5}13.0 & 110 & 85.7 \\
 & 1.7 & N & 268 & 0 & 183 (68.3\%) & 67 (25.0\%) & 17 (6.3\%) & 1 (0.4\%) & \cellcolor[HTML]{FBE4D5}\textbf{0.5} & 7.0 & 97 & 15.6 \\
Gemma-4 & 12 & Y & 358 & 0 & 300 (83.8\%) & 52 (14.5\%) & 5 (1.4\%) & 1 (0.3\%) & \textbf{2.5} & 3.7 & 165 & 269.7 \\
 & 12 & N & 372 & 0 & 285 (76.6\%) & 63 (16.9\%) & 23 (6.2\%) & 1 (0.3\%) & \textbf{3} & 4.8 & 169 & 67.2 \\
Gemma-3 & 12 & N & 330 & 0 & 235 (71.2\%) & 39 (11.8\%) & 55 (16.7\%) & 1 (0.3\%) & \cellcolor[HTML]{FBE4D5}\textbf{0} & 3.4 & 133 & 60.5 \\
 & 4 & N & 492 & 0 & 308 (62.6\%) & 134 (27.2\%) & 50 (10.2\%) & 0 (0.0\%) & \cellcolor[HTML]{FBE4D5}\textbf{0} & 2.8 & 205 & 36.2 \\
 & 1 & N & 803 & 14 & 329 (41.0\%) & 200 (24.9\%) & 209 (26.0\%) & 65 (8.1\%) & \cellcolor[HTML]{FBE4D5}\textbf{0.5} & 4.9 & 412 & 47.3 \\
Llama-3.2 {\tiny(8B: 3.1)} & 8 & N & 362 & 0 & 256 (70.7\%) & 58 (16.0\%) & 48 (13.3\%) & 0 (0.0\%) & \cellcolor[HTML]{FBE4D5}\textbf{0} & 3.8 & 167 & 35.7 \\
 & 3 & N & 395 & 0 & 222 (56.2\%) & 112 (28.4\%) & 52 (13.2\%) & 9 (2.3\%) & \cellcolor[HTML]{FBE4D5}\textbf{0.5} & 6.0 & 181 & 22.4 \\
 & 1 & N & 553 & 0 & 231 (41.8\%) & 225 (40.7\%) & 86 (15.6\%) & 11 (2.0\%) & \cellcolor[HTML]{FBE4D5}\textbf{2} & 7.8 & 254 & 13.5 \\
\addlinespace[3pt]
\multicolumn{13}{@{}l}{\textbf{ALL CASES (pooled)}} \\
Qwen3 & 14 & Y & 1405 & 0 & 1054 (75.0\%) & 132 (9.4\%) & 204 (14.5\%) & 15 (1.1\%) & \textbf{17.5} & \textbf{5.1} & 153 & 193.9 \\
 & 14 & N & 1037 & 0 & 788 (76.0\%) & 71 (6.8\%) & 151 (14.6\%) & 27 (2.6\%) & \textbf{20.5} & \textbf{3.9} & 150 & 74.6 \\
 & 8 & Y & 1409 & 0 & 1090 (77.4\%) & 131 (9.3\%) & 182 (12.9\%) & 6 (0.4\%) & \textbf{20} & \textbf{3.4} & 137 & 103.5 \\
 & 8 & N & 1062 & 0 & 801 (75.4\%) & 129 (12.1\%) & 129 (12.1\%) & 3 (0.3\%) & \textbf{19.5} & \textbf{3.6} & 139 & 37.5 \\
 & 4 & Y & 1566 & 0 & 1180 (75.4\%) & 209 (13.3\%) & 149 (9.5\%) & 28 (1.8\%) & \textbf{16.5} & \textbf{5.0} & 161 & 138.0 \\
 & 4 & N & 1404 & 0 & 1079 (76.9\%) & 174 (12.4\%) & 130 (9.3\%) & 21 (1.5\%) & \textbf{11.5} & \textbf{4.8} & 162 & 35.4 \\
 & 1.7 & Y & 1113 & 0 & 765 (68.7\%) & 239 (21.5\%) & 73 (6.6\%) & 36 (3.2\%) & \textbf{14} & \textbf{9.7} & 114 & 76.5 \\
 & 1.7 & N & 930 & 0 & 662 (71.2\%) & 179 (19.2\%) & 49 (5.3\%) & 40 (4.3\%) & \textbf{10} & \textbf{6.0} & 111 & 16.8 \\
Gemma-4 & 12 & Y & 1140 & 0 & 1004 (88.1\%) & 55 (4.8\%) & 80 (7.0\%) & 1 (0.1\%) & \textbf{22.5} & \textbf{3.3} & 161 & 270.9 \\
 & 12 & N & 1178 & 0 & 1012 (85.9\%) & 84 (7.1\%) & 81 (6.9\%) & 1 (0.1\%) & \textbf{23} & \textbf{3.6} & 164 & 68.5 \\
Gemma-3 & 12 & N & 1062 & 0 & 749 (70.5\%) & 134 (12.6\%) & 135 (12.7\%) & 44 (4.1\%) & \cellcolor[HTML]{FBE4D5}\textbf{4} & \textbf{5.4} & 136 & 61.2 \\
 & 4 & N & 1618 & 20 & 908 (56.1\%) & 314 (19.4\%) & 297 (18.4\%) & 99 (6.1\%) & \cellcolor[HTML]{FBE4D5}\textbf{1} & \textbf{5.3} & 257 & 43.9 \\
 & 1 & N & 2468 & 74 & 991 (40.2\%) & 552 (22.4\%) & 598 (24.2\%) & 327 (13.2\%) & \cellcolor[HTML]{FBE4D5}\textbf{1} & \textbf{6.8} & 401 & 48.2 \\
Llama-3.2 {\tiny(8B: 3.1)} & 8 & N & 1252 & 0 & 824 (65.8\%) & 158 (12.6\%) & 203 (16.2\%) & 67 (5.4\%) & \cellcolor[HTML]{FBE4D5}\textbf{2} & \textbf{5.3} & 186 & 42.6 \\
 & 3 & N & 1395 & 0 & 824 (59.1\%) & 272 (19.5\%) & 211 (15.1\%) & 88 (6.3\%) & \textbf{6.5} & \textbf{7.5} & 208 & 25.0 \\
 & 1 & N & 2056 & 1 & 800 (38.9\%) & 637 (31.0\%) & 315 (15.3\%) & 304 (14.8\%) & \cellcolor[HTML]{FBE4D5}\textbf{4}\textsuperscript{a} & \cellcolor[HTML]{FBE4D5}\textbf{12.6} & 290 & 16.1 \\
\bottomrule
\end{tabular*}}

\par\vskip4pt
\noindent\parbox{\linewidth}{\sffamily\scriptsize\raggedright
\textbf{B} = parameters (billions); \textbf{T} = thinking mode; \textbf{C} = labelled claims; \textbf{E} = ECHO lines (not claims, excluded). Percentages are shares of C. \textbf{TR} = Q3 trap resistance, sum over seeds of 1 (resisted), 0.5 (mitigated) or 0 (fell into the trap): out of 10 per case, 30 pooled; shaded green $\geq$ 80\%, red $\leq$ 20\%. \textbf{SI} = seed instability, mean of the four category-share SDs across seeds (percentage points); pooled row uses the degrees-of-freedom-weighted within-case SD; shaded $\geq$ 10. \textbf{AWC} = average words per response; \textbf{ART} = average response time (s); pooled AWC/ART are response-weighted means across cases. The Llama 8B model is Llama-3.1.\par
\textsuperscript{a} Paris Llama-3.2-1B: seeds 1 and 9 excluded (Q1 hit the 10,000-token limit in a degenerate loop); 8 seeds / 24 responses, so TR is out of 8 (pooled: 28).\par
}
\end{landscape}

\end{document}